\PassOptionsToPackage{linesnumbered,lined,boxed,commentsnumbered}{algorithm2e}

\documentclass[runningheads]{llncs}

\usepackage[british]{babel}

\usepackage[dvipsnames]{xcolor}
\usepackage{pgfplots}

\usepackage{amsfonts}
\usepackage{amsmath}
\usepackage{amssymb}

\usepackage{algorithm2e}
\usepackage{listings}

\usepackage{booktabs}

\usepackage{graphicx}
\usepackage{breakcites}
\usepackage[colorlinks=true]{hyperref}
\usepackage[nameinlink]{cleveref}
\hypersetup{colorlinks,breaklinks,citecolor=ForestGreen}

\begin{document}
\title{Random Projections with Best Confidence}
%
%
\author{Maciej Skorski}
\authorrunning{M. Skorski}
%
\institute{University of Luxembourg}
\maketitle              
\begin{abstract}%
  The seminal result of Johnson and Lindenstrauss on random embeddings 
  has been intensively studied in applied and theoretical computer science. Despite that vast body of literature, we still lack of complete understanding of statistical properties of random projections; a particularly intriguing question is: why are the theoretical bounds that far behind the empirically observed performance?
  
  Motivated by this question, this work develops Johnson-Lindenstrauss distributions with optimal, data-oblivious, statistical confidence bounds. These bounds are numerically best possible, for any given data dimension, embedding dimension, and distortion tolerance. They improve upon prior works in terms of statistical accuracy, as well as exactly determine the no-go regimes for data-oblivious approaches. Furthermore, the corresponding projection matrices are efficiently samplable.
  
  The construction relies on orthogonal matrices, and the proof uses certain elegant properties of the unit sphere. The following techniques introduced in this work are of independent interest: 
  a) a compact expression for distortion in terms of singular eigenvalues of the projection matrix, b) a parametrization linking the unit sphere and the Dirichlet distribution and c) anti-concentration bounds for the Dirichlet distribution.
  
  Besides the technical contribution, the paper presents applications and numerical evaluation along with working implementation in Python.
\end{abstract}


\section{Introduction}

\subsection{Background}

The seminal result of \cite{johnson1984extensions} on random embeddings 
is the cornerstone tool in dimension reduction. It rigorously shows that
euclidean distances are nearly preserved (low distortion) when
 high-dimensional data are projected into a lower-dimensional space using a random (appropriately sampled) matrix. What makes the random projections preferable in applications, over other dimension reductions techniques such as the principal component analysis or the singular value decomposition, are the speed, data-independence, and much stronger statistical guarantees (see~\cite{menon2007random,akselrod2011accelerating,vu2016random,bandeira2017compressive}).

The low-distortion property of random projections is very appealing
and makes them popular across many research areas. Among many applications one finds topics as diverse as
functional analysis (\cite{johnson2010johnson}), combinatorics (\cite{frankl1988johnson}), signal processing (\cite{haupt2006signal}),
proximity search~\cite{ailon2006approximate,indyk1998approximate},
low-rank matrix approximations (\cite{nguyen2009fast,clarkson2017low,ubaru2017low}), learning theory (\cite{arriaga2006algorithmic}), feature hashing (\cite{weinberger2009feature}), image hashing (\cite{lv2008fast}), classification (\cite{rahimi2007random,ghalib2020clustering}),
regression (\cite{maillard2012linear}), face recognition (\cite{goel2005face}), text mining (\cite{bingham2001random,lin2003dimensionality,ben2007analysis}), clustering (\cite{boutsidis2010random,tasoulis2014random,boutsidis2014randomized,makarychev2019performance,becchetti2019oblivious}), data storage (\cite{candes2008restricted,cormode2016stable}) and privacy (\cite{blocki2012johnson,kenthapadi2013privacy}).

Our focus is on the \emph{statistical guarantees} of the \emph{Distributional Johnson-Lindenstrauss Lemma}. The result states that
for every data dimension $m$ and the embedding dimension $n$ there exists a  random matrix $A$ of shape $n\times m$ (explicitly samplable), such that
for every non-zero data input $x\in\mathbb{R}^m$, and relative tolerance $0<\epsilon<\frac{1}{2}$, the euclidean norm is $\epsilon$-preserved with high confidence, provided that $n$ is sufficiently big. Formally:
\begin{align}\label{eq:djl}
 \mathbb{P}_{A}\left[  |\|A  x\|_2^2 -\|x\|_2^2|> \epsilon \|x\|_2^2\right] \leqslant  \mathrm{e}^{-\Omega(n\epsilon^2)}.
\end{align}
The goal of this work is to tackle the challenge of constructing random projections with the \emph{smallest possible distortion probability} defined as above, given data dimension $m$, embedding dimension $n$, and distortion $\epsilon$. 

\subsection{Related Work}

There have been several works on simplifying the proof and improving the provable confidence, namely: \cite{johnson1984extensions,frankl1988johnson,indyk1998approximate,achlioptas2003database,dasgupta2003elementary,matouvsek2008variants}. The best, up to date, upper bound for the distortion probability above is $2\exp\left(-\frac{n\epsilon^2}{4}\cdot\left(1-\frac{2\epsilon}{3}\right)\right)$~\cite{indyk1998approximate,achlioptas2003database} achieved for scaled Gaussian or Rademacher matrices.
; in other words the exponent is nearly $\frac{1}{4}$ for small distortions $\epsilon$. As for the impossibility results, we know that no distribution with $n<m/2$ can achieve distortion probability smaller than $\exp(-O(n\epsilon^2+1))$ for some unspecified constant ( \cite{alon2003problems,kane2011almost,jayram2013optimal}). Regarding this hidden constant, it has been recently shown in \cite{burr2018optimal} that it cannot be better, for any construction, than $\frac{1}{4}$ in some restricted asymptotic regimes, namely when $\epsilon\to0,n\epsilon^2\to+\infty$ and when $\frac{n}{m}\to0$. Interestingly, the evaluation experiments (see for example \cite{venkatasubramanian2011johnson,fedoruk2018dimensionality}), found theoretical guarantees far behind the observed performance.

The above discussion summarizes the state-of-art on confidence bounds, which is the subject of this paper. However, for readers interested in a broader scope of research on random projections, we would like to briefly discuss other lines of research. There are many works on trading the statistical accuracy  for certain algorithmic properties, such as sparsity and faster sampling (\cite{dasgupta2010sparse,ailon2013almost,kane2014sparser,cohen2018simple}) or specific matrix patterns (\cite{allen2014sparse,freksen2020using});
these properties can be somewhat improved under certain structural properties of datasets, if known in advance (\cite{bourgain2015toward}). Another trade-off is to extend the class of sampling distributions as much as possible (e.g. sub-gaussian matrices) as done by \cite{matouvsek2008variants,boucheron2003concentration}.

We also note that for certain datasets and for some parameter regimes, it is possible to slightly improve upon the DJL Lemma using
\emph{non-random} embeddings constructed combinatorically (\cite{nelson2014deterministic,larsen2016johnson}); these however are more of theoretical interests, and generally in theory and practice
DJL constructions are preferred, because of their \emph{data oblivious} properties (particularly useful for streaming, distributed and parallel computing).


\section{Results}

\subsection{Main Result: Characterizing Best Confidence}

We study the optimal error probability in \eqref{eq:djl}, given as the min-max program:
\begin{align}\label{eq:minmax}
\delta^{\mathrm{Best}}(m,n,\epsilon) = \inf_{\mathcal{A}\in \mathbb{P}(\mathbb{R}^{n\times m})}\sup_{x\in\mathbb{R}^m} \mathbb{P}_{A\sim \mathcal{A}}\left[  |\|A  x\|_2^2 -\|x\|_2^2|> \epsilon \|x\|_2^2 \right].
\end{align}
Since we maximize over the $m$-dimensional data inputs $x$, and minimize
over all the possible sampling distributions $\mathcal{A}$ for a projection from the dimension $m$ to $n$, this gives the best possible confidence bounds for the (oblivious) DJL Lemma. It may be convenient to think of this program as a two-player game: we seek for the best projection (minimizing over the distribution of the matrix), while the adversary controls the data and is seeking for the malicious input. 

Our main contribution shows that the ideal bound above is achievable,  characterizes it, and develops an explicit sampler.
Note that even the existence of a distribution achieving exactly (not approximately) the best bound is not trivial, because we deal with doubled optimization including distributions with unbounded support. As for the significance, our optimal confidence bound completes the line of research on improving the data-oblivious Distributional JL Lemma, establishing the numerically (not asymptotically) sharp no-go result.

We explain the  notation, before stating our result. 
By $\mathsf{Beta}(a,b)$ we denote the Beta distribution with shape parameters $a,b$; it has the cumulative distribution $B(z;a,b)/B(1;a,b)$ where
the incomplete Beta function is defined as 
$B(z;a,b)\triangleq \int_{0}^{z}z^{a-1}(1-z)^{b-1}\mbox{d}z$ (\cite[8.17]{NIST:DLMF}). 
By $\mathcal{O}(d)$ we denote the set of orthogonal matrices of shape $d\times d$ and by $I_{n,m}$ we denote the matrix of shape $n\times m$ with ones on the principal diagonal and zeros elsewhere (generalizing the identity matrix).

\begin{theorem}[Best Oblivious DJL Confidence]\label{thm:best_confidence_jl}
Let $1\leqslant n< m$ be integers, and $0<\epsilon < \frac{1}{2}$. Then the best value in \eqref{eq:minmax} is achievable and equals:
\begin{align}\label{eq:numeric}
\delta^{\mathrm{Best}}(m,n,\epsilon) = 1-\max_{\lambda}
\mathbb{P}\left[(1-\epsilon)\lambda\leqslant \mathsf{Beta}\left(\frac{n}{2},\frac{m-n}{2}\right)  \leqslant (1+\epsilon)\lambda\right].
\end{align}
Furthermore, let $\lambda$ be the maximizer of the right-hand side,
$U\sim \mathcal{O}(n)$, and $V\sim \mathcal{O}(m)$ be sampled uniformly and independently. Then
\begin{align}
    A^{\mathrm{Best}} = \lambda^{-1/2}\cdot U\cdot I_{n,m}\cdot V^T
\end{align}
is the random matrix which achieves the best value $\delta^{\mathrm{Best}}(m,n,\epsilon)$.
\end{theorem}
The prior results on confidence and our formula are summarized in \Cref{tab:bounds_history}.
\begin{table}[]
\resizebox{0.95\linewidth}{!}{
    \centering
    {\renewcommand{\arraystretch}{1.2}
    \begin{tabular}{c|c|c|c}
    \toprule
    Author & Error Upper Bound & Error Lower Bound & Restriction \\
    \midrule
    \cite{johnson1984extensions} & $2\exp(-\Omega(n\epsilon^2))$ & ? \\
    \cite{frankl1988johnson} & $2\exp\left(-\frac{n\epsilon^2}{9}(1+O(\epsilon))\right)$ & ? \\
     \cite{indyk1998approximate,achlioptas2003database} & $2\exp\left(-\frac{n\epsilon^2}{4}(1+O(\epsilon))\right)$ & ? \\
    \cite{alon2003problems} & & $\exp(-O(n\epsilon^2\log(\frac{1}{\epsilon})))$ & $n<\frac{m}{2}$ \\
    \cite{kane2011almost} & & $\exp(-O(n\epsilon^2))$ & $n<\frac{m}{2}$\\
        \cite{jayram2013optimal} & & $\exp(-O(n\epsilon^2))$ & $n<\frac{m}{2}$\\
        \cite{burr2018optimal} & & 
        $\exp\left(-\frac{n\epsilon^2}{4}(1+o(1))\right)$ & $\substack{n\epsilon^2\to\infty \\ \epsilon,\frac{n}{m}\to 0}$ \\
    \midrule
    \textbf{this work} & \multicolumn{2}{c}{$1-\max_{\lambda}
\mathbb{P}\left[(1-\epsilon)\lambda\leqslant \mathsf{Beta}\left(\frac{n}{2},\frac{m-n}{2}\right)  \leqslant (1+\epsilon)\lambda\right]$}  & None \\
    \bottomrule
    \end{tabular}
    }
}
    \caption{Confidence error $\delta^{\mathrm{Best}}(m,n,\epsilon)$ (Distortion Probability) in DJL.}
    \label{tab:bounds_history}
\end{table}

The step-by-step sampler construction is presented in \Cref{alg:sampler}.
\begin{algorithm}
\SetAlgoLined
\KwIn{Data dimension $m$, Embedding dimension $n$, Tolerance $\epsilon$} 
\KwResult{Best Oblivious DJL Matrix $A^{\mathrm{Best}}$}
\tcc{find the scaling factor}
\nl $\lambda \gets  \mathrm{argmax}_{\lambda}\mathbb{P}\left[(1-\epsilon)\lambda\leqslant \mathsf{Beta}\left(\frac{n}{2},\frac{m-n}{2}\right)  \leqslant (1+\epsilon)\lambda\right]$\\
\tcc{sample orthogonal matrices, uniformly and independently}
\nl $V\sim \mathcal{O}(m)$\\
\nl $U\sim \mathcal{O}(n)$\\
\tcc{build the projection }
\nl $A\gets \lambda^{-\frac{1}{2}} U I_{n,m} V^T$\\
\nl\Return{$A$}
 \caption{Best Sampler for Oblivious DJL Lemma.}
 \label{alg:sampler}
\end{algorithm}

\begin{remark}[Optimality under Worst vs Average Choice]
We define the best confidence in terms of a min-max problem, so that the construction is optimal under the \emph{worst choice} of the input. However, the proof actually establishes more, namely that the confidence cannot be improved even under the \emph{average choice from the unit sphere}, which is related to the use of Yao's Min-Max Principle.
\end{remark}

\begin{figure}[h]
    \centering
\begin{tikzpicture}[
declare function={ beta_pdf(\alpha,\beta,\x)                 = pow(\x,\alpha)*pow(1-\x,\beta)*1/factorial(\alpha-1)*1/factorial(\beta-1)*factorial(\alpha+\beta-1); 
                 },
]
\begin{axis}[domain=0.0:1.0,xlabel={$z$},ylabel={$\mathrm{pdf}(z)$}]
\addplot[draw=none, fill=gray!25,samples=100,domain=0.4995828:0.6106012]{beta_pdf(5,5,x)} \closedcycle;
\addplot[mark=none,thick,samples=100]{beta_pdf(5,5,x)};
\node (zplus) at (axis cs:0.6106012,0.0) {};
\draw (zplus) -- ++(45:350.0) node[above right] {$z(1+\epsilon)$};
\node (zminus) at (axis cs:0.4995828,0.0) {};
\draw (zminus) -- ++(135:350.0) node[above left] {$z(1-\epsilon)$};
\legend{$\mathsf{Beta}(a\,b)$};
\end{axis}
\end{tikzpicture}
\caption{The maximal statistical error equals the maximal area under the beta probability density, captured by an interval of form $[\lambda(1-\epsilon),\lambda(1+\epsilon)]$. The beta shape is $a=\frac{n}{2},b=\frac{m-n}{2}$.
In the picture $n=10,m=20$, the optimal value is $\lambda \approx \frac{5}{9}$ for small $\epsilon$.
}
\label{fig:optimization}
\end{figure}
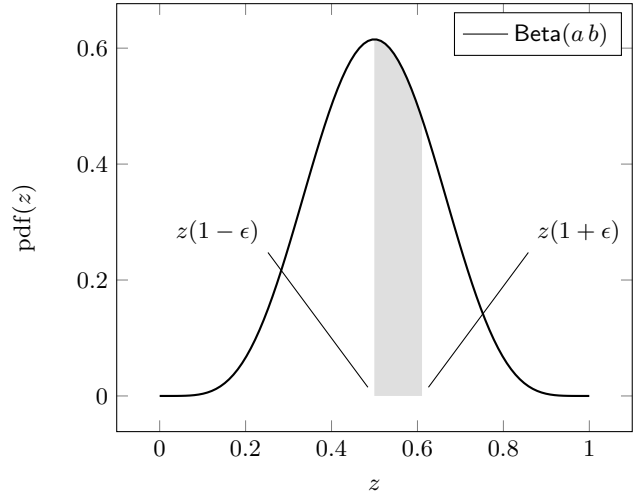

\begin{remark}[Parameter regimes]
Note that $\delta^{\mathrm{Best}}(m,n,\epsilon)=0$ for $n\geqslant m$ trivially (take the identity matrix as $A$); this regime is not interesting, as there is no dimension reduction. The interesting cases $n<m$ are fully addressed by the result; note that the optimal distribution depends on $m,n,\epsilon$.
\end{remark}
\begin{remark}[Construction] The sampler is built on appropriately scaled orthogonal matrices; there exist algorithms for efficiently sampling such matrices, see for example~\cite{10.2307/2156882,genz2000methods}. The scaling factor is chosen carefully as a solution to the one-dimensional numerical optimization problem \eqref{eq:numeric}; since the objective derivative can be explicitly calculated and the optimal point lies in the interval $\left(0,\frac{1}{1-\epsilon}\right)$, the program can be readily solved by modern data-science software, for example using \texttt{R} or \texttt{Python}. Below 
in \Cref{alg:implement} we demonstrate the implementation in the \texttt{SciPy} library for Python\footnote{The full implementation and all examples is available at \url{https://github.com/maciejskorski/confidence_optimal_random_embed}}. The optimization task can be best explained and interpreted geometrically, as visualized in \Cref{fig:optimization}.
\end{remark}

\begin{remark}[Closed-form Approximation]\label{rem:approximation}
We have exactly characterized best confidence. However, we mention
 the following, more readable and nearly sharp, convenient bounds:
\begin{align}\label{cor:closed_form_approximation}
\begin{split}
 \min\left\{ \mathbb{P}[B>(1+\epsilon)\mathbb{E}[B]] ,\
        \mathbb{P}[B<(1-\epsilon)\mathbb{E}[B]]\right\}
\leqslant \delta^{\mathrm{Best}}(m,n,\epsilon) \\
\delta^{\mathrm{Best}}(m,n,\epsilon)\leqslant
 \mathbb{P}[B>(1+\epsilon)\mathbb{E}[B]] +
        \mathbb{P}[B<(1-\epsilon)\mathbb{E}[B]], 
\end{split}
\end{align}
where we denote $B=\mathsf{Beta}\left(\frac{n}{2},\frac{m-n}{2}\right)$. The proof of this fact appears later, in the discussion of applications.
\end{remark}

\begin{lstlisting}[language=Python,frame=single,basicstyle=\scriptsize,keywordstyle=\bfseries,caption={Numerically Finding Optimal Confidence and Scaler \eqref{eq:numeric}},label={alg:implement}]
from scipy.optimize import minimize
from scipy.stats import beta

def optimal_djl(m,n,eps):
  ''' confidence-optinal sampler for DJL'''
  a = n/2
  b = (m-n)/2
  z0 = a/(a+b)
  dist = beta(a,b) 
  fun = lambda z:-dist.cdf((1+eps)*z)+dist.cdf((1-eps)*z)
  betainc_jac = lambda z: dist.pdf(z)
  jac = lambda z:-(1+eps)*betainc_jac((1+eps)*z)+(1-eps)*betainc_jac((1-eps)*z)
  out = minimize(fun,x0=z0,jac=jac,method='Newton-CG')
  scale,delta = out.x,1+out.fun
  return scale,delta
\end{lstlisting}

\subsection{Techniques of Independent Interest}

The proof of \Cref{thm:best_confidence_jl} builds on three elegant facts of broader interest. 
Below we abstract them as independent results and discuss in more detail.

\subsubsection{Explicit Distortion with Latent Singular Values}

Since the matrix product scales linearly with the input norm, the DJL Lemma reduces to the question about measure concentration on the unit sphere. In fact we know that random sphere points tend to be "hardest" (giving the worst confidence) for the DJL Lemma, as shown by an application of the Yao 
(\cite{kane2011almost,burr2018optimal}). 
The core of our approach is the observation that the distortion
on the random sphere point can be very conveniently expressed (as a diagonal quadratic form) in terms of the matrix singular eigenvalues. This is formally stated below.
\begin{theorem}\label{thm:distortion_singular_latent}
Let $X=(X_1,\ldots,X_m)$ be uniformly distributed on the unit sphere. Let $A$ be any $n\times m$
random matrix independent of $X$, where $n\leqslant m$, and $\lambda_1,\ldots,\lambda_n$ be the eigenvalues of $A A^T$. Then the following holds:
\begin{align}
 \|AX\|_2^2 \sim \sum_{k=1}^{n} \lambda_k X_k^2.
\end{align}
\end{theorem}
\begin{remark}
The matrix $A A^T$ is positive semi-definite (it is the so called Grammian matrix~\cite{deza1997geometry}), so the eigenvalues $\lambda_k$ are non-negative.
\end{remark}

\subsubsection{Sphere Sampling with Dirichlet Distribution}

To effectively handle calculations on the unit sphere, 
we develop the parametrization linking it to the \emph{Dirichlet Distribution}. This is a novelty 
in the context of other works that used complicated sphere paramaterizations, formally justified by calculus on differential forms (\cite{kane2011almost,burr2018optimal}), and in a wider context 
of sphere samplers, very important to Monte Carlo methods, as it is not addressed by extensive surveys
(\cite{Roberts2019}). In the theorem below, by $\mathsf{Dirichlet}(\alpha)$ we denote the Dirichlet distribution with the vector parameter $\alpha$.

\begin{figure}[h]
    \centering
    \includegraphics[scale=0.5]{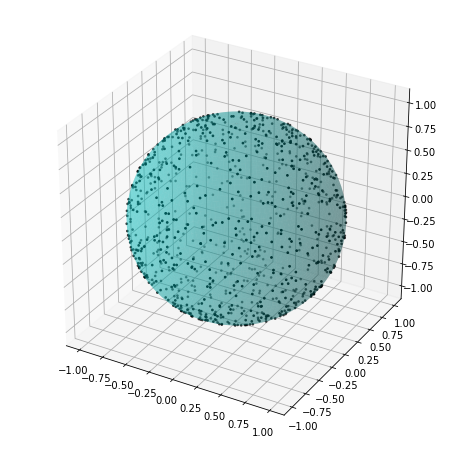}
    \caption{Sampling the unit sphere according to \Cref{thm:unit_sphere_dirichlet}, in $m=3$ dimensions.}
    \label{fig:sampling_sphere_dirichlet}
\end{figure}

\begin{theorem}\label{thm:unit_sphere_dirichlet}
Let $(X_1,\ldots,X_m)$ be uniform on the unit sphere in $\mathbb{R}^m$. Then:
\begin{align}
    (X_1^2,\ldots,X_m^2) \sim \mathsf{Dirichlet}\left(\frac{1}{2}\mathbf{1}_m \right)
\end{align}
where $\mathbf{1}_m$ denotes the vector of $m$ ones.
\end{theorem}
\begin{remark}
Let $Z_k=X_k^2$ and $(\varepsilon_k)_k$ be independent Rademacher variables (that is $\pm 1$ with equal probability). Then $(\varepsilon_k\sqrt{Z_k})_k$ is uniform on the unit sphere, as illustrated by the numerical simulation shown in \Cref{fig:sampling_sphere_dirichlet}.
\end{remark}

\subsubsection{Anti-concentration of Dirichlet Distribution}

The following result establishes \emph{sharp anti-concentration bounds} for weighted sums of components of Dirichlet's distribution. This is of broader interest  due to the popularity of Dirichlet distribution in statistics; such weighted sums appear in many applications (for a detailed discussion, see for example \cite{provost2000distribution}). 

\begin{theorem}\label{thm:dirichlet_anticoncentrate}
Let $Z=(Z_1,\ldots,Z_m)$ follow the Dirichlet distribution with parameters $(\alpha_1,\ldots,\alpha_m)$.
Let $W=(W_1,\ldots,W_m)$ be a vector of any non-negative random variables independent of $Z$.
Then for any non-empty and strict subset $I$ of $\{1\ldots m\}$ and real numbers $0<p<q$ the following holds:
\begin{align}
\min_{W}\mathbb{P}\left[ \sum_{k \in I} W_k Z_k\not\in [p,q] \right] =  
1-\max_{z>0} \left[p z\leqslant \mathsf{Beta}(a,b)\leqslant q z\right],
\end{align}
where $a=\sum_{i\in I} \alpha_i$ and $b=\sum_{i\not\in I}\alpha_i$, and $\mathsf{Beta}(a,b)$ is the Beta distribution.
\end{theorem}

\subsection{Applications}

\subsubsection{Dimension Estimation for Data Science Usage of Random Projections}

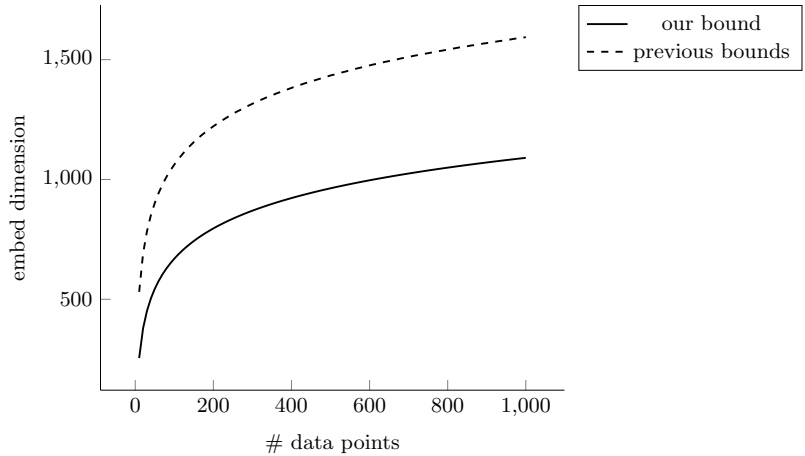
\begin{figure}[h]
\centering
\begin{tikzpicture}[scale=0.9]
\begin{axis}[
xlabel = {\# data points},
ylabel = {embed dimension},
legend pos=outer north east,
axis y line*=left,
axis x line*=bottom
]
\addplot[thick] table [x=n. points,y=my_dim,col sep=comma] {jl_dim.csv};
\addplot[thick,dashed] table [x=n. points,y=their_dim,col sep=comma] {jl_dim.csv};
\legend{our bound,previous bounds}
\end{axis}
\end{tikzpicture}
\caption{The minimal embedding dimension $n$ which guarantees distortion $\epsilon\leqslant 0.2$ for all pairwise distances of the given number of data points $x$.}
\label{fig:better_bounds}
\end{figure}

A good practice is to conservatively estimate the dimension before compressing the data with random projections. The DJL Lemma combined with a a union bound gives then provable guarantees. Such tests are implemented in 
modern data-science software, for example in the popular \texttt{Scikit-learn} library for Python (\cite{scikit-learn}). The problem with currently available bounds is that they
are based on overly conservative estimates from prior works, which creates the false impression that random projections should not be used. Our bounds give the more accurate answer, as illustrated in \Cref{fig:better_bounds} (the best previous bounds used for comparison are from \cite{indyk1998approximate,achlioptas2003database}). The Python code is available in \Cref{sec:better_bounds}.

\subsubsection{Dependency on Feature Dimension}

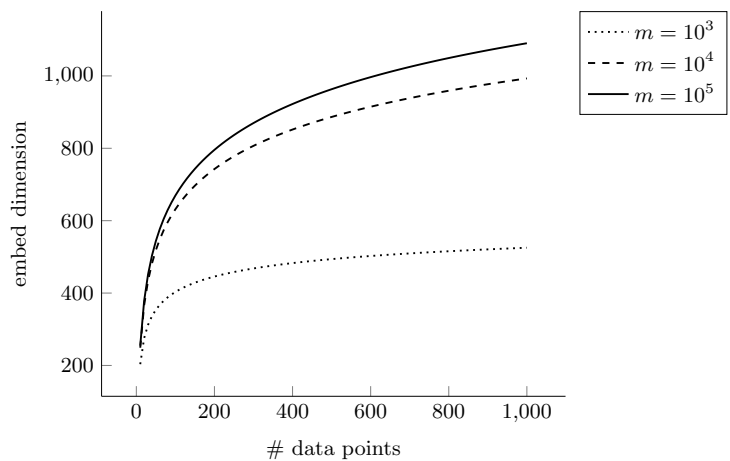
\begin{figure}
\centering
\begin{tikzpicture}[scale=0.9]
\begin{axis}[
xlabel = {\# data points},
ylabel = {embed dimension},
legend pos=outer north east,
axis y line*=left,
axis x line*=bottom
]
\addplot[thick,dotted] table [x=sample,y=1000,col sep=comma] {jl_feature_dim.csv};
\addplot[thick,dashed] table [x=sample,y=10000,col sep=comma] {jl_feature_dim.csv};
\addplot[thick,solid] table [x=sample,y=100000,col sep=comma] {jl_feature_dim.csv};
\legend{$m=10^3$,$m=10^4$,$m=10^5$}
\end{axis}
\end{tikzpicture}
\caption{The minimal embedding dimension $n$ which guarantees distortion $\epsilon\leqslant 0.2$ for all pairwise distances of the given number of data points $x$.
The results are better (that is, the embedding dimension is smaller) when the data dimension $m$ is smaller.}
\label{fig:feature_impact}
\end{figure}

Prior works have studied versions of DJL Lemma that are \emph{data-dimension} independent. However, even if we make no prior assumptions on the data structure, its dimension is known; is thus interesting to see the impact of the data dimension. This impact can be seen with the help of our optimal bounds; below in \Cref{fig:feature_impact} we show that knowing the data dimension helps improving the bound, with considerable impact when the dimension is of moderate magnitude.

\subsubsection{Closed-Form Upper Bounds on Confidence}

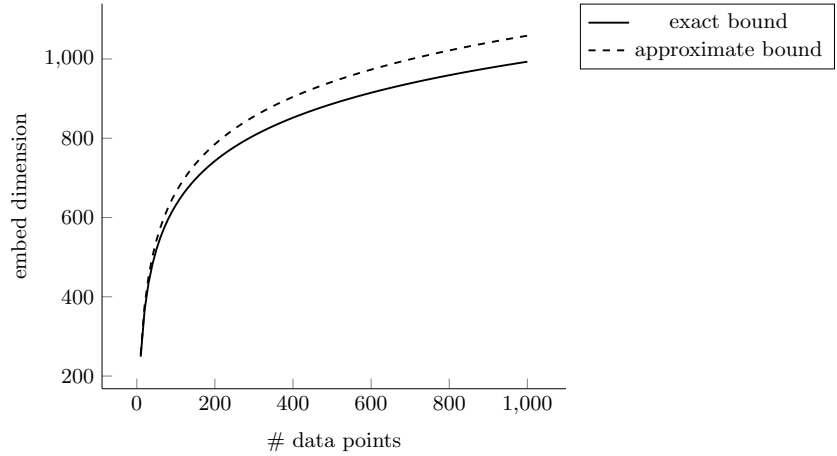
\begin{figure}[h]
\centering
\begin{tikzpicture}[scale=0.9]
\begin{axis}[
xlabel = {\# data points},
ylabel = {embed dimension},
legend pos=outer north east,
axis y line*=left,
axis x line*=bottom
]
\addplot[thick] table [x=n. points,y=exact,col sep=comma] {jl_approx_dim.csv};
\addplot[thick,dashed] table [x=n. points,y=approx,col sep=comma] {jl_approx_dim.csv};
\legend{exact bound,approximate bound}
\end{axis}
\end{tikzpicture}
\caption{The minimal embedding dimension $n$ with distortion $\epsilon\leqslant 0.2$ for all pairwise distances of the  data points: exact (\Cref{thm:best_confidence_jl}) and approximate (\Cref{rem:approximation}) bounds.}
\label{fig:approx_bounds}
\end{figure}

Let $B=\mathsf{Beta}\left(\frac{n}{2},\frac{m-n}{2}\right) $. Instead of optimizing numerically the bound in \Cref{thm:best_confidence_jl}, let us specialize $\lambda = \mathbb{E}[B]$. This way we obtain the following convenient upper-bound
\begin{align}
    \delta^{\mathrm{Best}}(m,n,\epsilon) \leqslant \Pr[|B-\mathbb{E}[B]|> \epsilon \mathbb{E}[B]],
\end{align}
which is very close to the optimal value, but
 does not involve optimization. The comparison of the approximate and exact bounds is given in \Cref{fig:approx_bounds}.

\subsubsection{Lower Bounds (Impossibility Results) on Confidence}
Again, let $B=\mathsf{Beta}\left(\frac{n}{2},\frac{m-n}{2}\right)$. Then $\delta^{\mathrm{Best}}(m,n,\epsilon) =\min_{\lambda}\Pr[B\not\in [\lambda(1-\epsilon),\lambda(1+\epsilon)]]$.
Further, $\Pr[B\not\in [\lambda(1-\epsilon),\lambda(1+\epsilon)]] = 
\Pr[B<(1-\epsilon)\lambda]+\Pr[B>(1+\epsilon)\lambda]$.
Considering that for any $\lambda$ it holds that either $\lambda \leqslant \mathbb{E}[B]$ or
$\lambda>\mathbb{E}[B]$, we obtain:
\begin{align}
        \delta^{\mathrm{Best}}(m,n,\epsilon) \geqslant 
        \min\left\{ \mathbb{P}[B>(1+\epsilon)\mathbb{E}[B]],
        \mathbb{P}[B<(1-\epsilon)\mathbb{E}[B]]\right\}.
\end{align}
This lower bound implies impossibility results obtained in prior works by 
\cite{kane2011almost} and \cite{burr2018optimal} (when combined with accurate approximations for tails of the beta distribution, such as those in (\cite{zhang2020non}).

Note that taking into account the previous upper bound, we prove \Cref{rem:approximation}.

\subsubsection{Use as Benchmark}

The fact that our result is numerically optimal for data-oblivious setup, and also easy to explicitly compute, makes it a perfect reference tool. When comparing
the theoretical and empirical performance (such as in works of \cite{venkatasubramanian2011johnson,fedoruk2018dimensionality}), our \Cref{thm:best_confidence_jl} now clarifies how big that gap actually is. Similarly for theoretical research, quantifying the best possible oblivious bound serves as a reference point for non-oblivious approaches, and also determines the range of possible improvements.

\section{Proofs}

\subsection{Proof of \Cref{thm:best_confidence_jl}}

Define 
\begin{align}
g(m,n,\epsilon)\triangleq 1-\max_{\lambda}
\mathbb{P}\left[\mathsf{Beta}\left(\frac{n}{2},\frac{m-n}{2}\right) \in \left[\lambda(1-\epsilon),\lambda(1+\epsilon)\right]\right].
\end{align}
It suffices to show that
\begin{align}\label{eq:inf_program}
    \inf_{\mathcal{A}\in\mathbb{P}(\mathbb{R}^{n\times m})}\ \sup_{x\in\mathbb{R}^m}\mathbb{P}_{A\sim\mathcal{A}}\left[  |\|A  x\|_2^2 -\|x\|_2^2| > \epsilon \|x\|_2^2 \right] = g(m,n,\epsilon).
\end{align}
We first prove that for $A= A^{\mathrm{Best}}$ sampled as described in \Cref{thm:best_confidence_jl}, and every fixed $m$-dimensional non-zero vector $x$ the following holds:
\begin{align}
\mathbb{P}_{A}\left[  |\|A  x\|_2^2 -\|x\|_2^2| > \epsilon \|x\|_2^2 \right] = g(m,n,\epsilon),
\end{align}
which in turn implies that the inequality "$\leqslant $" holds in \eqref{eq:inf_program}.

We observe that the condition under the probability is homogeneous (quadratic in $\|x\|_2$), thus we can restrict $\|x\|_2=1$. Now $x$ is on the unit sphere $\mathbb{S}^{m-1}$, and thus 
$X=V^T x$ is uniformly distributed on $\mathbb{S}^{m-1}$, because $V$ is orthogonal. We can write $Ax= A' X$ where
$A'=\lambda^{-\frac{1}{2}}\cdot U I_{m,n}$. Then $A'{A'}^T = \lambda^{-1}  U I_{n,m} I_{n,m}^T U^T  = \lambda^{-1} U I_{n,n} U^T = \lambda^{-1} U U^T = \lambda^{-1} I_{n,n}$, where we used the structure of $I_{n,m}$ and the orthogonality of $U$. Now by \Cref{thm:distortion_singular_latent} we obtain:
\begin{align}
    \|Ax\|_2^2 \sim \lambda^{-1} \sum_{k=1}^{n} X_k^2.
\end{align}
Combining this with \Cref{thm:unit_sphere_dirichlet}, for $Z\sim\mathsf{Dirichlet}\left(\frac{1}{2}\mathbf{1}_m\right)$ we obtain
that for every non-zero $x$:
\begin{align}
\mathbb{P}_{A}\left[  |\|A  x\|_2^2 -\|x\|_2^2|> \epsilon \|x\|_2^2 \right] =
1-\mathbb{P}\left[1-\epsilon \leqslant \lambda^{-1} \sum_{k=1}^{n}Z_k \leqslant 1+\epsilon\right].
\end{align}
By the properties of the Dirichlet distribution (\cite{albert2012dirichlet}) 
we have that $\sum_{k=1}^{n}Z_k \sim \mathsf{Beta}(a,b)$
with $a = \frac{n}{2}$ and $b=\frac{m-n}{2}$. Thus, by the definition of $\lambda$, we conclude that the right-side equals $ g(m,m,\epsilon)$.

In the second part we show that for every \emph{fixed} matrix $A$ we have:
\begin{align}
\mathbb{P}_{x\sim \mathbb{S}^{m-1}}\left[  |\|A  x\|_2^2 -\|x\|_2^2| > \epsilon \|x\|_2^2 \right] \geqslant g(m,n,\epsilon),
\end{align}
where $\mathbb{S}^{m-1}$ denotes unit sphere in $m$-dimensions; this establishes the inequality "$\geqslant $" in \eqref{eq:inf_program}
(by replacing the expectation over $x$ with the maximum and taking the expectation over the distribution of $A$). 
Let $X$ be uniform on $\mathbb{S}^{m-1}$. Let $\lambda_1,\ldots,\lambda_n$ be the eigenvalues of $A A^T$ (they are deterministic numbers).
Then:
\begin{align}
     \|AX\|_2^2 \sim  \sum_{k=1}^{n} \lambda_k X_k^2.
\end{align}
Combining this with \Cref{thm:unit_sphere_dirichlet}, for $Z\sim\mathsf{Dirichlet}\left(\frac{1}{2}\mathbf{1}_m\right)$ we obtain:
\begin{align}
\begin{split}
    \mathbb{P}_{x\sim \mathbb{S}^{m-1}}\left[  |\|A  x\|_2^2 -\|x\|_2^2|> \epsilon \|x\|_2^2 \right] & =
1-\mathbb{P}\left[1-\epsilon \leqslant  \sum_{k=1}^{n}\lambda_k Z_k \leqslant 1+\epsilon\right] \\
& = \mathbb{P}\left[\sum_{k=1}^{n}\lambda_k Z_k \not\in [1-\epsilon, 1+\epsilon]\right] .
\end{split}
\end{align}
Regardless of the choice of $\lambda_k$, by \Cref{thm:dirichlet_anticoncentrate} we get the lower bound:
\begin{align}
\mathbb{P}_{x\sim \mathbb{S}^{m-1}}\left[  |\|A  x\|_2^2 -\|x\|_2^2|> \epsilon \|x\|_2^2 \right] \geqslant
1-\max_{\lambda}\mathbb{P}\left[\lambda(1-\epsilon) \leqslant  B \leqslant \lambda(1+\epsilon)\right],
\end{align}
where $B=\mathsf{Beta}\left(\frac{n}{2},\frac{m-n}{2}\right)$. This completes the proof, since the expression on the right-hand side equals $g(m,n,\epsilon)$.

\subsection{Proof of \Cref{thm:unit_sphere_dirichlet}}

The Dirichlet distribution $Z=(Z_1,\ldots,Z_m)$ with parameters $\frac{1}{2}\mathbf{1}_m$
can be sampled as $Z_k = \Gamma_k / \sum_{i=1}^{n}\Gamma_i$,
where $\Gamma_i$ are independent and follow the Gamma distribution with the parameters: shape $a=\frac{1}{2}$ and the rate $b=1$ (\cite{albert2012dirichlet}).
We next observe that $\Gamma_i \sim \frac{1}{2}\chi_1$, 
where $\chi_1$ is the chi-squared distribution with 1 degree of freedom (\cite{thom1958note}).
By definition, $\chi_1\sim N^2$ where $N$ is the standard normal random variable. Therefore, we obtain:
\begin{align}
    Z_k \sim \frac{\frac{1}{2}N_k^2}{\sum_{i=1}^{m}\frac{1}{2}N_i^2} = \frac{N_k^2}{\sum_{i=1}^{m}N_i^2},\quad N_i \sim^{iid} \mathsf{Norm}(0,1).
\end{align}
We now recall that the normalized normal vector generates the uniform measure on the sphere (see \cite{muller1959note,marsaglia1972choosing}.
); more precisely if $(X_1,\ldots,X_m)$ is the uniform distribution on the sphere, then
\begin{align}
\left(  \frac{N_k}{\sqrt{\sum_{i=1}^{m}N_i^2}}\right)_{k=1}^{m} \sim (X_k)_{k=1}^{m},
\end{align}
and combining this with the previous equation we get
\begin{align}
    (Z_k)_{k=1}^{m} \sim (X_k^2)_{k=1}^{m},
\end{align}
so the result follows.

\subsection{Proof of \Cref{thm:distortion_singular_latent}}

By the SVD decomposition (see~\cite{stewart2001matrix}) we have $A = U \Sigma V^T$ where $U,V$ are orthogonal with shapes 
$n\times n$ and $m\times m$ respectively, and $\Sigma $ is an $n\times m$ diagonal (rectangular) matrix with 
real values $\sigma_1,\ldots,\sigma_n$ on the principal diagonal (recall that $n\leqslant m$).
Using the orthogonality of $U$ and $V$ we obtain:
\begin{align}
    \begin{split}
    \|AX\|_2^2 &= \|U \Sigma V^T X\|_2^2 \\
    & = \|\Sigma V^T X\|_2^2. 
    \end{split}
\end{align}
Since $V$ is orthogonal, so is $V^T$. Since $X$ is uniform on the unit sphere and independent of $V$ and $\Sigma$, we see that $V^T X$ conditioned on the pair $\Sigma,V$ is also uniform on the unit sphere and thus distributed as $X$:
\begin{align}
    V^T X | \Sigma,V \sim X,
\end{align}
as the sphere uniform measure is invariant under orthogonal transforms. 
Combining the two equations above we express the squared distance $\|AX\|_2$, conditioned on $\Sigma,V$ as follows:
\begin{align}
\left. \|AX\|_2^2 \right| \Sigma,V \sim \|\Sigma X\|_2^2,
\end{align}
and since the right-hand side does not depend on $V$, this gives:
\begin{align}
\|AX\|_2^2 \sim \|\Sigma X\|_2^2 = \sum_{k=1}^{n}\sigma_k^2 X_k^2.
\end{align}
It remains to observe that 
\begin{align}
    \mathrm{diag}(\sigma_1,\ldots,\sigma_n)^2 = \Sigma\Sigma^T,
\end{align}
and, because $V$ and $U$ are orthogonal, that:
\begin{align}
    A A^T = U\Sigma \Sigma^T U^T = U\Sigma\Sigma^T U^{-1}.
\end{align}
Thus, we see that $\lambda_k=\sigma_k^2$ are eigenvalues of $A A^T$.

\subsection{Proof of \Cref{thm:dirichlet_anticoncentrate}}

Let $Z=(Z_1,\ldots,Z_m)$ follow the Dirichlet distribution with parameters $(\alpha_1,\ldots,\alpha_m)$. 
Let $I\subset \{1\ldots m\}$ and $J=\{1\ldots m\}\setminus I$; by the self-normalizing properties (see \cite{albert2012dirichlet}):
\begin{align}
\left.    \frac{1}{\sum_{i\in I}Z_i} (Z_i)_{i\in I} \right| (Z_i)_{i\in J} \sim \mathsf{Dirichlet}((\alpha_i)_{i\in I}).
\end{align}
Since $\sum_{i=1}^{m}Z_i = 1$, we have that $\sum_{i\in I}Z_i = 1-\sum_{i\in J}Z_i$
depends only on the components $(Z_i)_{i\in J}$. Therefore, from the above identity we obtain: 
\begin{align}
\left.\frac{1}{\sum_{i\in I}Z_i}(Z_i)_{i\in I} \right|\sum_{i\in I}Z_i \sim \mathsf{Dirichlet}((\alpha_i)_{i\in I}).
\end{align}
Further, by the proportion properties~\cite{albert2012dirichlet}:
\begin{align}
   \sum_{i\in I}Z_i \sim \mathsf{Beta}\left(\sum_{i\in I} \alpha_i,\sum_{i\in J} \alpha_i\right).
\end{align}
Let $B,D$ be such that $D\sim \mathsf{Dirichlet}((\alpha_i)_{i\in I})$,
 $B\sim\mathsf{Beta}\left(\sum_{i\in I} \alpha_i,\sum_{i\in J} \alpha_i\right)$ and that random variables $B,D,Z,W$ where $W=(W_1,\ldots,W_m)$ are independent (this is possible, since $Z,W$ are independent). By the two equations above :
\begin{align}
    (Z_i)_{i\in I} \sim B\cdot D.
\end{align}
Now, for any \emph{deterministic} scalar vector $w=(w_i)_{i\in I}$ it holds that:
\begin{align}
    \sum_{i\in I} w_i Z_i \sim B\cdot \sum_{i\in I}w_i D_i.
\end{align}
Since $(W_i)_{
i \in I}$ is independent of $B,D,Z$ we obtain:
\begin{align}
    \sum_{i\in I} W_i Z_i\sim B\cdot \sum_{i\in I}W_i D_i.
\end{align}
Consider now any fixed numbers $0<p<q$. Denote $U = \frac{1}{\sum_{i\in I}W_i D_i}$, then:
\begin{align}
\mathbb{P}\left[\sum_{i\in I} W_i Z_i \in [p,q]\right] = \mathbb{P}_{B,U}\left[p U \leqslant B \leqslant q U\right].
\end{align}
Since $B,U$ are independent, we have that
$\mathbb{P}_{B,U}\left[p U \leqslant B \leqslant q U\right]
= \mathbb{E}_{u\sim U}\mathbb{P}_B\left[p u \leqslant B \leqslant q u\right]$ and thus
\begin{align}
\begin{split}
\mathbb{P}\left[\sum_{i\in I} W_i Z_i \in [p,q]\right] & \leqslant  \max_{u>0}\mathbb{P}[p u \leqslant B \leqslant q u ],
\end{split}
\end{align}
with the equality when $\mathbb{P}[U=u^{*}]=1$ where
\begin{align}
    u^{*}=\mathrm{argmax}_{u}\mathbb{P}_{B}\left[p u \leqslant B \leqslant q u 
     \right].
\end{align}
The upper bound is indeed achieved with the following choice of $W$:
\begin{align}
\forall i\in I:\    W_i = \frac{1}{\cdot u^{*}},
\end{align}
because then $U = \frac{1}{\sum_{i\in I} D_i/u^{*}}=u^{*}$, we use $\sum_{i\in I} D_i = 1$. Thus, we have shown
\begin{align}
    \max_{W}\mathbb{P}\left[\sum_{i\in I} W_i Z_i \in [p,q]\right] =  \max_{u>0}\mathbb{P}[p u \leqslant B \leqslant q u ].
\end{align}
The result follows now by noticing that
$\min_{W}\mathbb{P}\left[\sum_{i\in I} W_i Z_i \not\in [p,q]\right]=1-\max_{W}\mathbb{P}\left[\sum_{i\in I} W_i Z_i \in [p,q]\right]$.

\section{Conclusion}

This work constructed the \emph{confidence-optimal} Distributional Johnson-Lindenstrauss distribution; the optimal bounds and the sampler are built based on the solution of a 1-dimensional optimization program involving the Beta distribution. With best bounds clearly established, the only way to improve further is by non-oblivious bounds.

In our approach the critical role play the techniques for handling the distortion probability on the unit sphere, the sphere parametrization using the Dirichlet distribution, and anticoncentration inequalities for the Dirichlet distribution, which we stated are of independent interest.


\bibliographystyle{apalike}
\bibliography{citations}

\appendix

\section{}\label{sec:better_bounds}

\begin{lstlisting}[language=Python,frame=single,basicstyle=\ttfamily\footnotesize,caption={Minimal Dimension Bounds for JL Lemma},label={alg:implement_min_dim}]
import numpy as np
from scipy.optimize import bisect
from sklearn.random_projection import johnson_lindenstrauss_min_dim
from matplotlib import pyplot as plt

def optimal_djl_dim(N,m,eps):
  ''' optimal dimension with many data points '''
  fun = lambda n: optimal_djl(m,n,eps)[1]-2/(N*(N-1))
  return bisect(fun,10,m-10)

m = 1e5
eps = 0.2
sample = np.linspace(10,1000,100)
my_dim = [optimal_djl_dim(N,m,eps) for N in sample]
plt.plot(sample,my_dim,label='our bound')
their_dim = [johnson_lindenstrauss_min_dim(N,eps) for N in sample]
plt.plot(sample,their_dim,label='previous bound')
plt.legend()
plt.show()
\end{lstlisting}

\begin{lstlisting}[language=Python,frame=single,basicstyle=\ttfamily\footnotesize,caption={Impact of Data Dimension for JL Lemma},label={alg:implement_dim_impact}]
eps = 0.2
sample = np.linspace(10,1000,100)

my_dims = []

for m in [1e3,1e4,1e5]:
  my_dim = [optimal_djl_dim(N,m,eps) for N in sample] 
  my_dims.append(my_dim)
  plt.plot(sample,my_dim,label='%s'%m)
plt.legend()
plt.show()
\end{lstlisting}

\begin{lstlisting}[language=Python,frame=single,basicstyle=\ttfamily\footnotesize,caption={Sampling Sphere with Dirichlet Distribution},label={alg:implement_sphere_sample}]
from matplotlib import pyplot as plt
from mpl_toolkits.mplot3d import axes3d
from scipy import stats

def sample_dirichlet(size=1,dim=3):
  w = stats.dirichlet(0.5*np.ones(dim)).rvs(size=size)
  sign = np.random.choice([-1,1],size=(size,dim),p=[0.5,0.5])
  return sign*w**0.5

fig = plt.figure(figsize=(8,8))
ax = fig.add_subplot(111, projection='3d')

u = np.linspace(0, 2 * np.pi, 120)
v = np.linspace(0, np.pi, 60)
x = np.outer(np.cos(u), np.sin(v))
y = np.outer(np.sin(u), np.sin(v))
z = np.outer(np.ones(np.size(u)), np.cos(v))
ax.plot_surface(x, y, z,  rstride=1, cstride=1, color='c', alpha = 0.3, linewidth = 0)

xi, yi, zi = sample_dirichlet(1000).T
ax.scatter(xi, yi, zi,color="k",s=3)
\end{lstlisting}

\end{document}